\documentclass[hidelinks]{article}

\usepackage{arxiv}
\usepackage{setspace}
\usepackage[utf8]{inputenc} % allow utf-8 input
\usepackage[T1]{fontenc}    % use 8-bit T1 fonts
\usepackage{hyperref}       % hyperlinks
\usepackage{url}            % simple URL typesetting
\usepackage{booktabs}       % professional-quality tables
\usepackage{amsfonts}       % blackboard math symbols
\usepackage{amssymb}
\usepackage{amsthm}
\usepackage{amsmath}
\usepackage{mathrsfs}
\usepackage{dsfont}
\usepackage{nicefrac}       % compact symbols for 1/2, etc.
\usepackage{microtype}      % microtypography
\usepackage{graphicx}
\usepackage[natbibapa]{apacite}
\newtheorem{definition}{Definition}

\title{Pruned Wasserstein Index Generation Model and \textit{wigpy} Package}

\author{
 Fangzhou Xie \\
  Department of Economics, New York University\\
	Department of Economics, Rutgers University\\
  \url{fangzhou.xie@{nyu,rutgers}.edu} \\
}

\begin{document}
\maketitle
\begin{abstract}
	Recent proposal of Wasserstein Index Generation model (WIG) has shown a new direction
	for automatically generating indices. However, it is challenging in practice to fit large
	datasets for two reasons. First, the Sinkhorn distance is notoriously expensive to compute
	and suffers from dimensionality severely. Second, it requires to compute a full $N\times N$ matrix
	to be fit into memory, where $N$ is the dimension of vocabulary. When the dimensionality is too
	large, it is even impossible to compute at all. I hereby propose a Lasso-based shrinkage method
	to reduce dimensionality for the vocabulary as a pre-processing step prior to fitting the WIG model.
	After we get the word embedding from Word2Vec model, we could cluster these high-dimensional vectors
	by $k$-means clustering, and pick most frequent tokens within each cluster to form the
	“base vocabulary”. Non-base tokens are then regressed on the vectors of base token to
	get a transformation weight and we
	could thus represent the whole vocabulary by only the “base tokens”. This variant,
	called pruned WIG (pWIG), will enable us to shrink vocabulary dimension at will
	but could still achieve high accuracy.
	I also provide a \textit{wigpy}\footnote{
		\href{https://github.com/mark-fangzhou-xie/wigpy}{https://github.com/mark-fangzhou-xie/wigpy}}
	module in Python to carry out computation in both flavor.
	Application to Economic Policy Uncertainty (EPU) index is showcased as comparison with existing
	methods of generating time-series sentiment indices.
\end{abstract}

% keywords can be removed

\section{Introduction}
Recently, the Wasserstein Index Generation model \citet{xie2020} was proposed to
generate time-series sentiment indices automatically. There have been several
methods
\citep{azqueta-gavaldon2017,baker2016,castelnuovo2017,ghirelli2019}
% (AzquetaGavaldón, 2017; Baker, Bloom, & Davis, 2016; Castelnuovo & Tran, 2017; Ghirelli, Pérez, & Urtasun, 2019)
proposed to generate time series sentiment indices, but,
to the best of my knowledge, WIG is the first automatic method to produce
sentiment indices completely free of manual work.

The WIG model runs as follows. Given a set of documents, each of which is
associated with a time-stamp, it will first cluster them into several topics,
shrink each topic to a sentiment score, then multiply weights for each document
to get document sentiment, and then aggreagate over each time period.
However, its computation on large dataset come with two challenges:
(1) the calculation for Sinkhorn algorithm suffers from its notoriously
computational complexity and the computation will soon become prohibitive;
(2) this Optimal Transport-based method requires to compute a full \(N\times N\)
matrix, where \(N\) is the size of vocabulary, and it will become impossible to fit
this distance matrix into memory after some threshold. Therefore,
I propose a pruned Wasserstein Index Generation model (pWIG) to reduce
dimensionality of vocabulary prior to fitting into the WIG model.
This variant could represent the whole corpus in a much smaller vocabulary
and then be fit in any memory-limited machine for the generation of time-series
index. What is more, I also provide the wigpy \footnote{
	Codes are available here:
	\href{https://github.com/mark-fangzhou-xie/wigpy}
	{https://github.com/mark-fangzhou-xie/wigpy}.}
package for Python that could
perform both version of WIG computation1.

This paper first contributes to the EPU literature and tries to
provide better estimations of that seminal time-series indices automatically.
This article also relates itself to the new area of Narrative Economics
\citet{shiller2017}, where we could extract time-series sentiment indices
from textual data, and thus provide a better understanding of how do
narratives and sentiments relate to our economy.

\section{Pruned Wasserstein Index Generation Model}
We first review the original WIG model.

\subsection{Review of Wassserstein Index Generation model}

A major component of WIG model is the Wasserstein Dictionary Learning
\citep{schmitz2018}.
Given a set of document $Y=\left[y_{m}\right] \in \mathbb{R}^{N \times M}$,
each doc $y_{m} \in \Sigma^{N}$ is associated with a timestamp and
$N, M$ are length of dictionary and number of documents in corpus,
respectively. Our first step is to cluster documents into topics
$T=\left[t_{k}\right] \in \mathbb{R}^{N \times K},$ where $K \ll M,$
and associated weights
$\Lambda=\left[\lambda_{m}\right] \in \mathbb{R}^{K \times M}.$
Thus, for a single document $y_{m},$ we could represent it as
$y_{m} \approx t_{k} \lambda_{m} .$ Documents and topics lie in
$N$-dimensional simplex, and are word distributions.
Another important quantity for computing WIG,
is the cost matrix
$C^{N \times N}$ and $C_{i j}=d^{2}\left(x_{i}, x_{j}\right),$
where each $x_{i} \in \mathbb{R}^{1 \times D}$ is the
$D$-dimensional word embedding vector for the $i$ -th word in the vocabulary.
In other words, matrix $C$ measures the ``cost'' of moving masses of words,
and now we can proceed and define the Sinkhorn Distance.

\begin{definition}[Sinkhorn Distance]
	Given discrete distributions $\mu, v \in \mathbb{R}_{+}^{N},$ and $C$ as cost matrix,
	$$
		\begin{array}{c}
			S_{\varepsilon}(\mu, v ; C):=\min _{\pi \in \mathbb{R}(\mu, v)}\langle\pi, C\rangle+\varepsilon \mathcal{H}(\pi) \\
			\text {s.t. } \Pi(\mu, v):=\left\{\pi \in \mathbb{R}_{+}^{N \times N}, \pi 1_{N}=\mu, \pi^{T} 1_{N}=v\right\}
		\end{array}
	$$
	where $\mathcal{H}(\pi):=\sum_{i j} \pi_{i j} \log \left(\pi_{i j}-1\right),$ negative entropy, and $\varepsilon$ is the Sinkhorn regularization weight.
\end{definition}

We could then set up the loss function and minimization problem as follows:

\begin{equation}
	\begin{aligned}
		 & \min \sum_{m=1}^{M} \mathcal{L}\left(y_{m}, y_{S_{\varepsilon}}\left(T(R), \lambda_{m}(A) ; C, \varepsilon\right)\right)                                  \\
		 & s.t.~t_{n k}(R):=\frac{e^{r_{n k}}}{\sum_{n^{\prime}} e^{r_{n^{\prime}} k}}, \lambda_{n k}(A):=\frac{e^{a_{k m}}}{\sum_{k^{\prime}} e^{a_{k^{\prime} m}}}
	\end{aligned}
\end{equation}

By this formula, we wish to minimize the divergence between original
document $y_{m}$ and the predicted (reconstructed)
$y_{\mathcal{S}_{\mathcal{E}}}(.)$ given by Sinkhorn distance.
Moreover, the constaints of this minimization problem considers
Softmax operation on each of the columns of the matrices $R$ and $A$,
so that $T$ and $\Lambda$ will be (column-wise) discrete densities,
as is required by the Sinkhorn distance.

For computation, we first initialize matrices $R$ and $A$ by drawing from
Standard Normal distribution and then perform Softmax to obtain $T$ and $\Lambda .$
During training process, we keep track of computational graph and obtain the
gradient $\nabla_{T} \mathcal{L}(\cdot ; \varepsilon)$ and
$\nabla_{\Lambda} \mathcal{L}(\cdot ; \varepsilon)$ with respect
to $T$ and $\Lambda.$ $R$ and $A$ are then optimized by Adam optimizer \citep{kingma2015}
after each batch,
and the automatic differentiation is done by PyTorch framework \citep{paszke2017}.

After conducting WDL on documents for clustering, the next step of WIG
would be to generate time-series indices based on the topics.
The model first reduce each topic vector $t_{k}$ to a scalar by
SVD and then multiply the weight matrix to get document-wise sentiment
score for the whole corpus.
We then add up the scores for each month and then produce the final monthly index.

\subsection{Pruned WIG (pWIG) Model}

Although enjoying many nice theoretical properties \citep{villani2003},
the computation for Optimal Transport has been known for its complexity.
This burden has been eased by \citet{cuturi2013} and it has attracted
much attention in machine learning community since then.

However, there are still two aspects that hindering our application to
textual analysis. First of all, vocabulary will easily go to a very
large one, and the computation for Sinkhorn loss will soon become
prohibitive. Moreover, after passing a certain point, it not even
possible to fit the distance matrix $C$ into the memory,
especially when considering the limited VRAM for GPU acceleration.\footnote{
	My configuration is Nvidia 1070Ti (8G). Under single presicion,
	each digit will occupy 4 byte and in my case, I can only fit,
	theoretically at most, a square matrix of dimension 44,721.
	I have a relative small dataset from The New York Times and
	my vocabulary is of length 9437, but many NLP applications
	will have much more tokens than I do. In such a case,
	the WIG model will become infeasible.}

I therefore propose the following procedure to reduce the vocabulary
dimension and could avoid feeding the full vocabulary matrix into WIG model.
It first clusters all word vectors by $k$-means clustering, and then selects
a subset of tokens from each of the cluster to form ``base tokens.''\footnote{
	The number of tokens to be considered as “base tokens” is arbitrary, meaning that the compression ratio could potentially be made arbitrarily small. In other words, the researcher could choose such a number that the model can be fitted into the memory of her machine, regardless of the number of tokens she had for the corpus. And that is exactly the way why we need to compress the dictionary by “pruning” some non-important tokens.
}
We could then use Lasso\footnote{
	A similar approach \citep{mallapragada2010} was proposed using group-LASSO to prune visual vocabulary, but in the area of image processing.}
to regress word vectors of all other tokens on
the vectors of these ``base tokens'' to ensure sparse weight vector,
which will have zero component on non-import features.

Formally speaking, we set up the following minimization problem for the $k$ -means clustering:

$$
	\operatorname{argmin}_{\mathcal{X}_{1}, \mathcal{X}_{2}, \cdots, \mathcal{K}_{n}} \sum_{k=1}^{K} \sum_{\mathcal{X} \in \mathcal{X}_{k}}\left\|x-\mu_{k}\right\|
$$

where $\mu_{k}$ is the mean of points in cluster $\mathcal{K}_{k}$ and $k \in\{1, \cdots, K\} .$ We can certainly choose some most frequent tokens from each cluster to form a final subset whose length matches our desire. \footnote{
	A very simple choice would be Word per Cluster $=\frac{\text { Maximum Vocabulary Length }}{\text { Number of Clusters }}$}
By doing so, we also represent the whole vocabulary by the most
representative tokens. The indices for these ``base tokens''
are collected in the index set,
$$
	\mathfrak{B}=\left\{b \in\{1, \cdots, N\} | x_{b}=1\right\}
$$

Obviously, $\mathfrak{B}^{C}$ is also defined by excluding ``base tokens''
from the whole vocabulary. $N$ is the size of vocabulary and $x_{b}$ is the
$b$-th token in the vocabulary.
Denote word vector for ``base tokens'' as $v_{b}$ and others as $v_{o},$
we have

$$
	\hat{\alpha}^{\text {lasso }}=\underset{\alpha}{\operatorname{argmin}}\left\{\frac{1}{2} \sum_{b=1}^{B}\left(v_{o}-\sum_{j=1}^{p} v_{b} \alpha_{o, b}\right)^{2}+\lambda \sum_{j=1}^{p}\left|\alpha_{o, b}\right|\right\}, \forall o
$$
For each $o,$ we will have a vector $\alpha_{o, b}$ of length $B,$
where $B$ is the dimension of ``base vocabulary''.

Previously in the WIG model, we obtain the word distribution for each single
document $y_{m}$ by calculating its word frequency, and that will give us
a $N$ -dimensional distribution vector. Here, in the pWIG variant,
we replace the non-base tokens by weighted base-tokens and could thus
represent the word simplex of documents in only $B$ -dimensional spaces.

Now that we have successfully represent our dataset in $s$ smaller vocabulary,
we could proceed to define our distance matrix
$C_{i j}=d^{2}\left(x_{i}, x_{j}\right),$ where $i, j \in \mathfrak{B} .$
Here we have everything we need for the regular WIG model and we fit it
using the shrinkage transformed word distributions and distance matrix.

\section{Numerical Experiments}

\subsection{\textit{wigpy} Package for Python}

To carry out the computation of WIG and pWIG model, I also provide the
\textit{wigpy} package under MIT license.
Notice that the WIG model in \textit{wigpy} package is a new implementation,
though part of the codes are modified from the codes of original WIG paper.

The main model is wrapped in the ``WIG'' class, where it contains a set
of hyper-parameters \footnote{
	For example, embedding depth ($emsize$), batch size ($batch\_size$),
	number of topics ($num\_topics$), sinkhorn regularization weight
	($reg$),
	optimizer learning rate ($lr$)
	(\(L2\) penalty for optimizer ($wdecay$),
	L1/LASSO weight ($l1\_reg$),
	maximum number of tokens allowed by pWIG algorithm ($prune\_topk$).}
to tune the model, and some parameters to control the behavior of
preprocessing and Word2Vec training process.

Notice that the previous implementation of WIG model only supports
hand-written Adam optimizer, and the optimization for document weights
were optimized column-wise. In other words, each document will only be
used to update the column of weight in matrix $\Lambda$ for that given
document. The new implementation wraps the whole model in PyTorch,
providing many optimizers to choose by PyTorch optimizer class.
What is more, each document will accumulate gradient and the whole
$\Lambda$ matrix will be updated all together.

\subsection{Application to Generating Economic Policy Uncertainty Index (EPU)}
To test for the pWIG model's performance, I run the model on the same dataset
from the WIG paper. It consists of news headlines collected from The New York
Times from 1980 to 2018. As I am implementing a new version of WIG,
as provided by the \textit{wigpy} module, I run the original WIG model
and report its result as well.

I run both variants of WIG model separately, by calling \textit{wigpy} package,
to set for hyperparameters by splitting training, evaluation, and
testing data as $60 \%, 10 \%,$ and $30 \%$ respectively.

For the original WIG, hyper-parameters are chosen as follows: depth of
embedding $D=50$ batch size $s=32,$ number of topics $K=4,$ learning rate for
Adam $\rho=0.001,$ Sinkhorn regularization weight $\varepsilon=0.1 ;$
for the pWIG, depth of embedding $D=50,$ batch size $s=$ 64,
number of topics $K=4,$ learning rate for Adam $\rho=0.001,$
Sinkhorn regularization weight $\varepsilon=0.08$
I also report Pearson's and Spearman's correlation test on four set of
automatically generated EPU indices
(one LDA-based EPU \citep{azqueta-gavaldon2017},
one WIG-based EPU \citep{xie2020}, and two flavor of WIG given by \textit{wigpy}
package in this paper), against the original EPU \footnote{
	\href{https://www.policyuncertainty.com/}{https://www.policyuncertainty.com/}.
}
\citep{baker2016}.

\begin{table}
	\centering
	\begin{tabular}{ccc}
		\hline
		EPU flavor & Pearson's        & Spearman's       \\
		\hline
		LDA        & 77.48\%          & 75.42\%          \\
		\hline
		WIG        & 80.24\%          & 77.49\%          \\
		\hline
		WIG-wigpy  & \textbf{80.53\%} & \textbf{77.71}\% \\
		\hline
		pWIG-wigpy & 80.50\%          & 77.64\%          \\
		\hline
	\end{tabular}
	\caption{Pearson’s and Spearman’s correlation statistics\protect\footnotemark}
	\label{tab:correlation}
\end{table}
\footnotetext{Since the LDA-based EPU was only available from 1989-2016,
	the test is performed using time-series indices within the same range.}

Apparently, as is shown in Table $1,$ all three WIG methods outperforms
LDA-based method by $3 \%$ in Pearson's test and more than $2 \%$ in
Spearman's test. This fact has been established by the previous WIG paper.
Moreover, if we compare results within three WIG-related methods, this new
implementation of original WIG in \textit{wigpy} package shows better result than
the previous implementation. The pruning method does not differ much from
the new implemented WIG algorithm, and is even better than the
previous implementation of original WIG!

\begin{table}
	\centering
	\begin{tabular}{ccccc}
		\hline
		           & VIX Pearson's & VIX Spearman's & Michigan Pearson's & Michigan Spearman's \\
		WIG-wigpy  & 34.20\%       & 19.56\%        & -56.40\%           & -49.38\%            \\
		pWIG-wigpy & 34.27\%       & 19.82\%        & -56.45\%           & -49.62\%            \\
	\end{tabular}
	\label{tab:2}
\end{table}

In Table 2, the correlation statistics between EPU generated by WIGs and two
other indices: VIX and Michigan Consumer Confidence Sentiment index.
As reported \citep{baker2016}, EPU has a correlation of 0.58 between
VIX and -0.742 between Michigan index. Since our objective is to produce a
similar index of EPU, but using an automatic approach, we should expect our
WIG-based EPU to have a similar relationship with these other two indices.
This is indeed the case here, and we can certainly observe the positive and
negative relationship when comparing the VIX and Michigan indices\footnote{
	It may be confusing why the “sentiment index” generated by WIG models has a
	negative relationship with “Michigan Consumer Sentiment index,” since both
	names contain “sentiment.” However, there is a clear distinction of the
	usage of the same word in two different contexts. The famous Michigan index
	is expressed as the consumer confidence levels, and the higher the index,
	the more confident the consumers are. The word “sentiment”, as used by WIG,
	is to capture the subjective information expressed in the texts. In the
	application of EPU, it is used to capture the intensity of opinions towards
	the uncertainty of policy, as conveyed by newspaper articles. It is very
	obvious that what it captures is negative feelings, and the higher the index,
	the more uncertain that people feel. In other words, although bearing the same
	word “sentiment” in their names, the underlying element is strikingly different
	and thus show a negative relationship between each other. Moreover, the WIG
	model does not limit its use in EPU. As soon as we apply the WIG models to
	other (textual) datasets, the meaning of “sentiment” will be changed accordingly.
	In total, the word “sentiment” used in WIG models is more versatile and should
	be distinguished from the usage as in the Michigan index.
}.

% \begin{figure}%[H]
% 	\centering
% 	\includegraphics[width=.8\linewidth]{epu.png}
% 	\caption{EPU indices given by different methods.}
% 	\label{fig:epu}
% \end{figure}

\section{Conclusion}
This paper further extends the Wasserstein Index Generation (WIG) model,
by selecting a subset of tokens to represent the whole vocabulary to
shrink the dimension. The showcase of generating EPU has shown that
the performance is retained while dimension being reduced. Moreover,
a package, \textit{wigpy}, is provided to carry out the computation
of two variants of WIG\@.

\newpage
% \bibliography{/Users/xiefangzhou/Library.bib}

%%% Comment out this section when you \bibliography{references} is enabled.

\end{document}